\documentclass[letterpaper, 10 pt, conference]{ieeeconf}  

\IEEEoverridecommandlockouts                              
\usepackage{graphicx}
\usepackage{morefloats}
\usepackage{subfig}

\title{\LARGE \bf
Extreme Augmentation : Can deep learning based medical image segmentation be trained using a single manually delineated scan?
}

\author{Bilwaj Gaonkar$^{1}$, Matthew Edwards$^{1}$, Alex Bui $^{2}$, Matthew Brown $^{3}$, Luke Macyszyn$^{1}$
\thanks{$^{1}$B. Gaonkar and L. Macyszyn are with the Department of Neurosurgery,
        University of California, Los Angeles
        {\tt\small bilwaj@gmail.com, LMacyszyn@mednet.ucla.edu}
        $^{2}$Alex Bui is the director of the Medical Imaging Informatics Group at UCLA Radiology
        $^{3}$Matthew Brown is the director of the Center for Computer Vision and Imaging Biomarkers (CVIB) 
        }%
}

\begin{document}

\maketitle
\thispagestyle{empty}
\pagestyle{empty}

\begin{abstract}
Yes, it can. Data augmentation is perhaps the oldest preprocessing step in computer vision literature. Almost every computer vision model trained on imaging data uses some form of augmentation. In this paper, we use the inter-vertebral disk segmentation task alongside a deep residual U-Net as the learning model, to explore the effectiveness of augmentation. In the extreme, we observed that a model trained on patches extracted from \textit{just one} scan, with each patch augmented 50 times; achieved a  Dice score of 0.73 in a validation set of 40 cases. Qualitative evaluation indicated a clinically usable segmentation algorithm, which appropriately segments regions of interest, alongside limited false positive specks. When the initial patches are extracted from nine scans the average Dice coefficient jumps to 0.86 and most of the false positives disappear. While this still falls short of state-of-the-art deep learning based segmentation of discs reported in literature, qualitative examination reveals that it does yield segmentation, which can be amended by expert clinicians with minimal effort to generate additional data for training improved deep models. Extreme augmentation of training data, should thus be construed as a strategy for training deep learning based algorithms, when very little manually annotated data is available to work with. Models trained with extreme augmentation can then be used to accelerate the generation of manually labelled data. Hence, we show that extreme augmentation can be a valuable tool in addressing scaling up small imaging data sets to address medical image segmentation tasks. 
\end{abstract}

\section{INTRODUCTION}
Medical image segmentation involves the automated or semi-automated delineation of physician specified anatomy using patient scans. Deep learning based algorithms for segmentation are considered state-of-the-art in this problem domain \cite{han2018spine,Jamaludin2016,Jamaludin2017,Whitehead2018,Moran2018,Lootus2015,Zheng2017}. However, it is unclear, just how many human annotated scans should be used to initially train a given algorithm. Human expert annotation is both cumbersome and expensive in the medical setting  \cite{Zheng2017,Gaonkar2017}. Thus, methods which can train with relatively fewer manually annotated data and yet achieve robust performance on segmentation tasks are especially valuable in the medical image analysis setting \cite{wachinger2018deepnat,shaikhina2017handling}. 

In this work we show that copious data augmentation can be applied to small training data sets to extract substantial performance gains for deep learning based segmentation. Data augmentation has been at the heart of improving the performance of machine learning models, especially in vision, from the earliest days\cite{witten2016data,Razavian2014,Sekuboyina2017,Jamaludin2016a,Kazemi2014}. Landmark papers have relied on some form of data augmentation as a key pre-processing step. \cite{krizhevsky2012imagenet,He2017}. Yet, data augmentation remains under appreciated, especially given how powerful it can be,   when data set sizes are small, like in medical image analysis.  This study is our attempt to highlight the importance of this key pre-processing step and document it's effectiveness.

\begin{figure}
  \centering
  \includegraphics[width=0.3\linewidth]{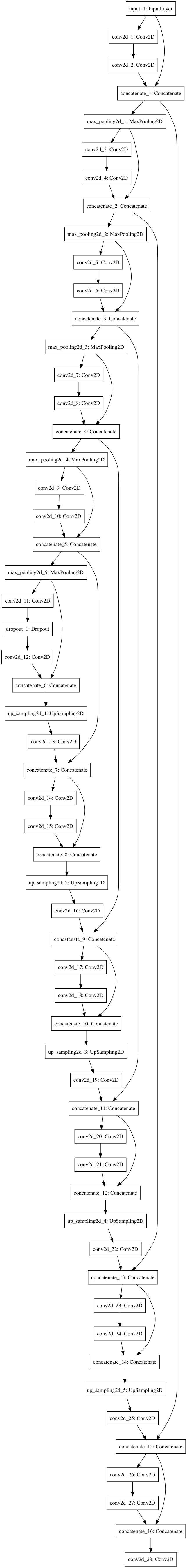}
  \caption{Residual U-Net model used in our experiments}
  \label{fig:model}
\end{figure}

We use a deep residual U-Net model \cite{ResUNet} shown in figure \ref{fig:model} for training. This model is derived from the older U-Net model \cite{Ronneberger:2015gk} which has been incredibly successful in medical image segmentation. In our study intervertebral disk segmentations were performed by medical staff under the supervision of an attending neurosurgeon (LM). Fifty scans, each belonging to a separate individual, obtained from 8 different scanners were used for the study. 10 scans were chosen randomly to be used for creating training datasets of varying sizes. The remaining 40 scans were used as a validation set.

We trained the deep residual U-Net model with varying degrees (5x,10x,20x,30x and 50x) of data augmentation, and by using either 1, 3, 5 or 9 scans for training purposes, and documented the results. The rest of this manuscript is organized into four sections. In the `Methods' section, we describe the architecture chosen and precisely detail the augmentation procedure used. In the `Results' section, we present graphical and tabular evidence to support our claims. In the conclusions section, we present our insights based on the results. In the discussion section, we discuss the advantages of being able to train with small datasets. 

\begin{figure}
  \centering
  \includegraphics[width=0.9\linewidth]{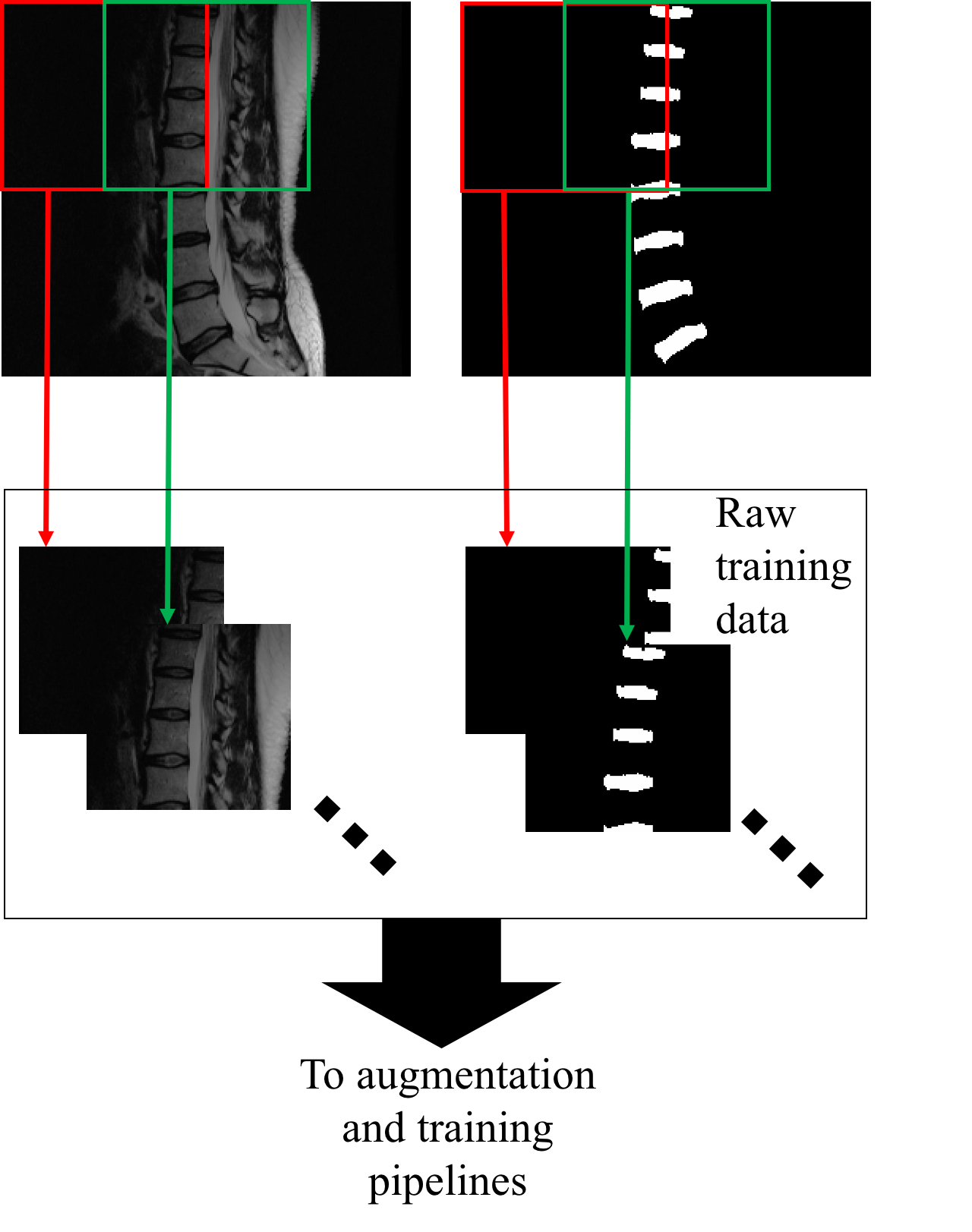}
  \caption{Patch generation schema used in our experiments}
  \label{fig:patches}
\end{figure}

\section{Methods}

\subsection{Model Architecture}
Figure 1 describes the fully convolutional model architecture used in our experiments. Briefly, there are 5 -residual downsampling stages wherein the number of filters doubles with every stage. There's a central residual stage and 5 up-sampling residual stages. Feed-forward connections between downsampling and upsampling stages follow the standard architectural paradigm of the U-Net.

The input layer was designed to ingest a single channel 128x128 px input image. The output layer was designed to generate a 128x128px segmentation map. All convolutional layers used ReLU activation \cite{Nair:2010vq} with 3x3 convolution filtering and a stride of 1.  The output layer used the sigmoid activation to ensure that the output stays bounded in the unit interval. Exact details of the network architecture are presented in Figure 1. The model was trained to maximize the Dice measure of overlap \cite{Zou2004,Warfield2004} between model output and segmentations in training data.

We implemented this network in the Keras \cite{chollet2015keras} package which is a deep learning API built on top of Google's Tensorflow \cite{abadi2016tensorflow} library.

\subsection{Data collection and pre-processing}

 The data used as a part of this work was obtained by  querying the University of California at Los Angeles (UCLA) picture archiving and communication system (PACS) for individuals who had undergone any form of spine imaging using the corresponding CPT (Current Procedural Terminology) codes \cite{Terminology1970} corresponding to lumbar MRI. The search yielded a large number of accession numbers, of which we randomly selected 50 for the purposes of experiments detailed here. This data was obtained under the IRB no. 16-000196.  
 
 All 50 images were downloaded from PACS, anonymized and resampled in the axial plane to 256x256-px. Subsequently each image was converted to the NIFTI \cite{Li:2016kj,Larobina2014} format and linearly histogram matched linearly to a template image using the SimpleITK \cite{lowekamp2013design} package. The template image intensities were originally scaled to lie between a maximum of 1 and a minimum of 0. Linear histogram matching, ensured that the same was true of each of the 50 images used in this study.
 
 Manual segmentation of inter vertebral disks was performed by two medical students using ITK-SNAP \cite{Yushkevich:2006fk} and validated by a resident physician. These manual segmentation data were used as ground truth.

\subsection{Patch set generation}
The central aim of the investigation presented here was to demonstrate that data augmentation, can be applied to solve challenging medical image segmentation problems even in the absence of large quantities of training data. Large annotated data sets are typically thought of as prerequisites for training deep learning methods. \cite{greenspan2016guest} In this work we show that this is not necessarily always the case. 

Recall that our network was designed to operate on 128x128 pixel patches of imaging data.  In our experiments we generate image patches from axial slices extracted from 3D data  using 64 px strides as depicted in figure 2. Input patches are collected from pre-processed input scan(s) and output patches are collected from corresponding manual segmentation(s). Perhaps a completely different model and patch generation scheme could have been used. However, our aim in this work is not to focus on model optimization, but rather to highlight the effectiveness of data augmentation. Hence, we used a particular instance of model architecture and patch generation schema.
\begin{figure}[h!]
  \centering
  \includegraphics[width=\linewidth]{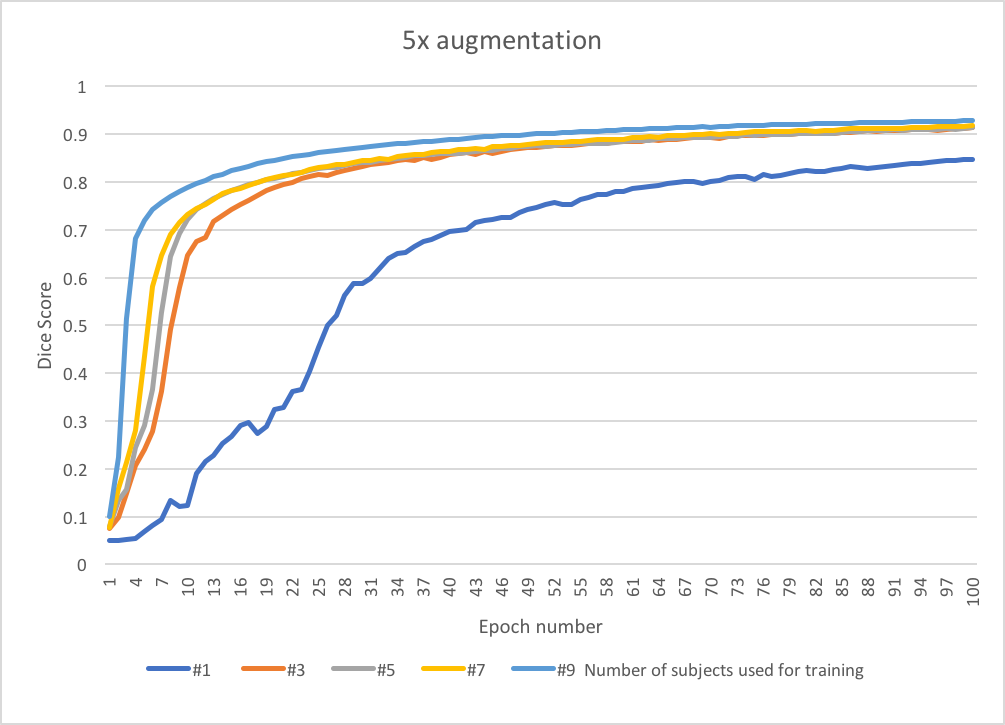}
  \includegraphics[width=\linewidth]{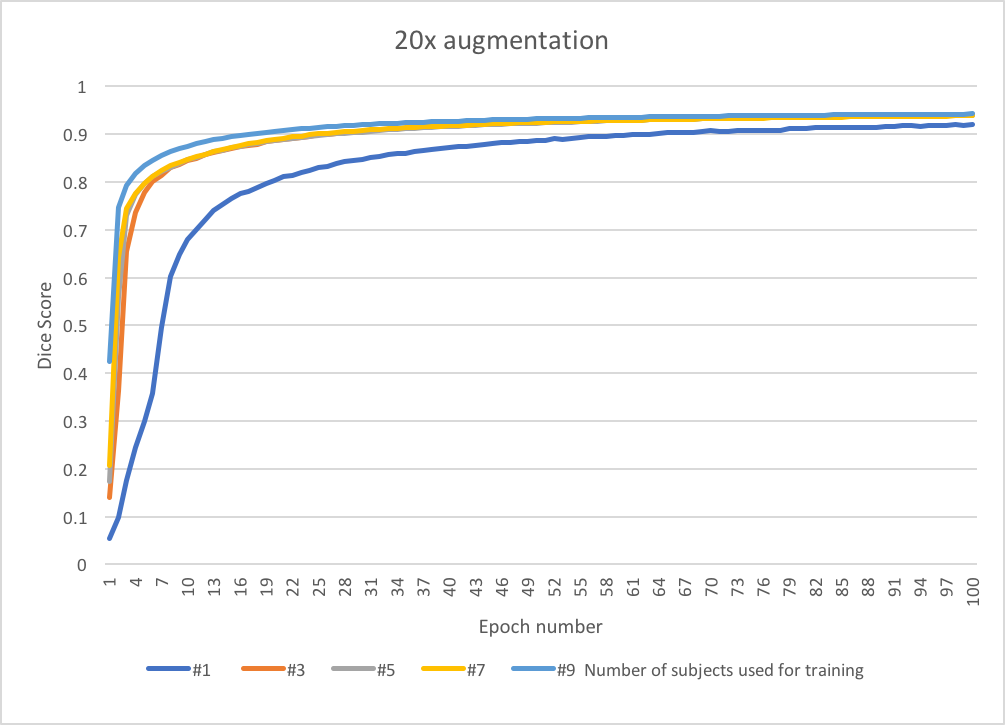}
  \includegraphics[width=\linewidth]{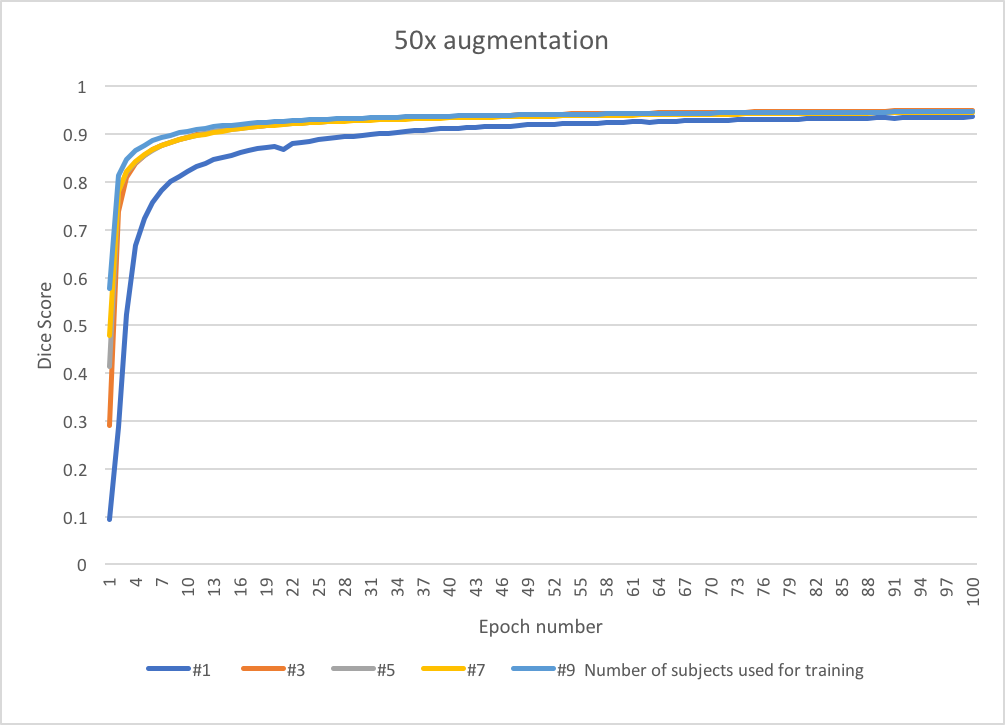}
  \caption{Training Dice scores curves grouped by augmentation levels, color coded by training data size.}
  \label{fig:auxtrain}
\end{figure}

\subsection{Augmentation}
Augmentation was performed on a patch by patch basis. It was performed by transforming each patch (and the corresponding segmentation) by some randomly picked combination of a translation, rotation and scaling. For each slice, the augmentation algorithm randomly picked an angle between $+/- 20^o$, a scaling factor between $[0.8,1.2]$
and x-translation and y-translation limited by $+/-50\;$ px.  Spine images tend to come to our PACS system in a specific orientation. Hence, flipping was not included in the augmentation heuristic used for the experiments presented here. 

To study the effect of augmentation, it is necessary to quantify the degree of augmentation applied to a training dataset. \textit{For the purposes of this paper augmentation levels were defined as follows. If each patch was 'augmented' using 5 randomly picked augmentation transformations, we defined the resulting post-augmentation dataset to be at the `5x' level of augmentation. We generated training data at 5x, 10x, 20x, 30x, 40x and 50x augmentation levels. }

\subsection{Experiment design and validation}
In our experiments, we trained the residual U-Net model on training data sets of various different sizes (1 scan, 3 scans , 5 scans, 7 scans and 9 scans). For each training data set size, we applied augmentation at all levels and recorded results for the first 100 epochs during training. Each trained model, obtained at the end of 100 epochs was used to segment disks using the 40 MRIs of the validation set.

\section{Results}
\subsection{Training curves}

\begin{figure}[h!]
  \centering
  \includegraphics[width=\linewidth]{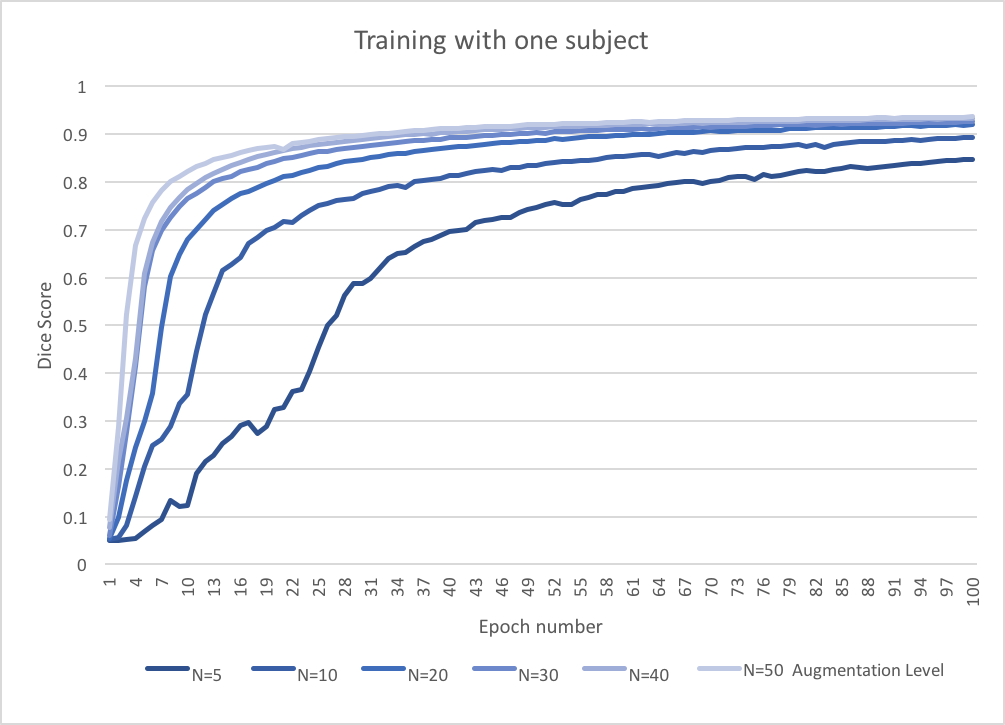}
  \includegraphics[width=\linewidth]{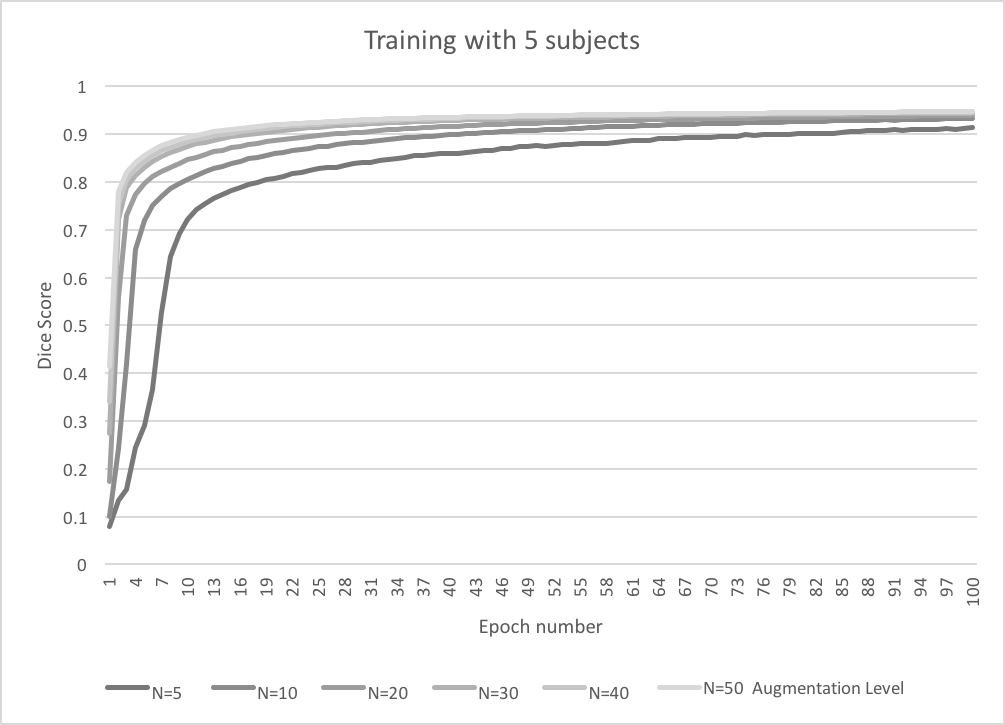}
  \includegraphics[width=\linewidth]{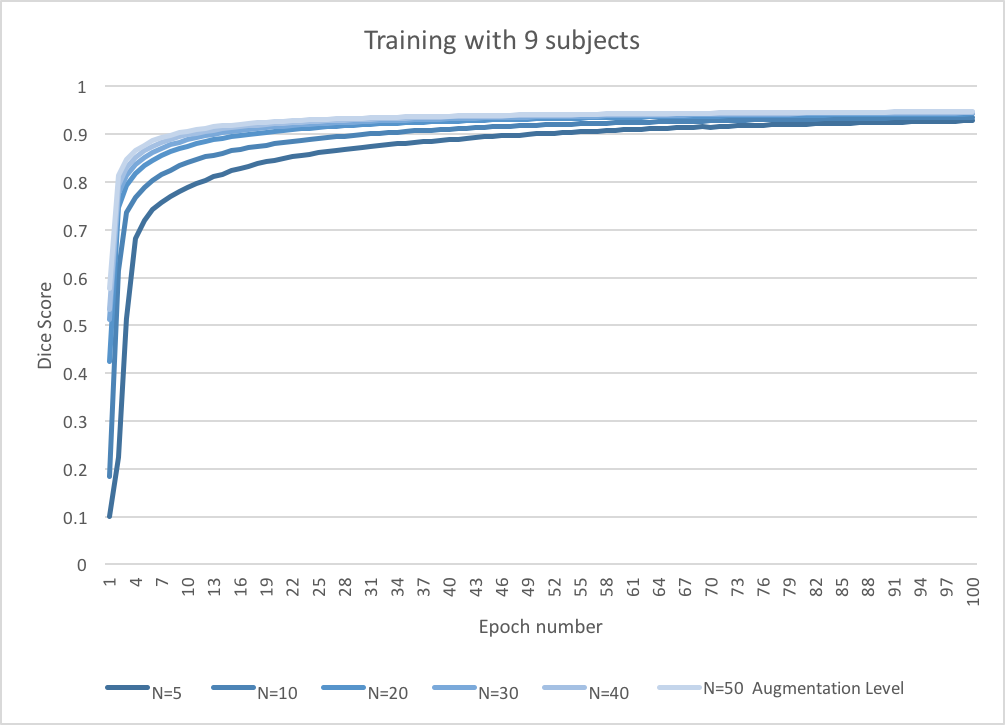}
  \caption{Training Dice scores curves grouped by training data set size. N represents augmentation level. For example N =50 represents 50x augmentation)}
  \label{fig:train}
\end{figure}

We recorded training accuracy in terms of the Dice score for the 100 epochs for models trained using training datasets of all sizes and at all augmentation levels. Table 1  presents the final training accuracy for each model at the 100th epoch. Note that a total of 30 models were trained in the course of the experiment, one model corresponding to each possible combination of the level of augmentation and the size of training data. 
  Training accuracy curves are presented in figures 3 and 4. Each plot in figure 3 groups training Dice scores across various sizes of training data but at a single augmentation level into a single plot. Only, data from models trained ta 5x, 20x and 50x levels are presented for the sake of brevity.
Each plot in figure 4 groups training curves across various augmentation levels, but belonging to particular size of training data into a single plot. Again due to lack of space, only curves corresponding to models trained using 1 scan, 5 scans and 9 scans are presented. 
\begin{table}
\centering
\caption{Training accuracy, as measured by Dice Score at 100th epoch }
\begin{tabular}{|l|l|l|l|l|l|} 
\hline
\# Training subjects ➡️    & 1                    & 3                    & 5                    & 7                    & 9      \\ 
\hline
Augmentation Level⬇️       &                      &                      &                      &                      &        \\ 
\hline
5x                         & 0.848                & 0.913                & 0.913                & 0.917                & 0.928  \\
10x                        & 0.892                & 0.930                & 0.932                & 0.931                & 0.936  \\
20x                        & 0.920                & 0.941                & 0.939                & 0.938                & 0.942  \\
30x                        & 0.926                & 0.946                & 0.944                & 0.942                & 0.944  \\
40x                        & 0.932                & 0.947                & 0.946                & 0.944                & 0.946  \\
50x                        & 0.936                & 0.949                & 0.947                & 0.945                & 0.947  \\
\hline
\end{tabular}
\end{table}

\subsection{Validation curves}
We used 40 scans for validation, none of which were in the training set  to generate figures 6 and 7. Each of the 30 models trained, were used to segment the validation cohort of 40 cases, and Dice scores comparing algorithmic segmentations to human generated segmentations were recorded. Mean Dice scores for the entire validation datasets, generated by each model are presented in Table 2. To give the reader a qualitative assessment of how powerful extreme augmentation is, we show in figure 5, a mid-sagittal slice with segmentation overlays generated by models at 50x data augmentation. Dice scores, at augmentation levels 5x, 20x and 50x, for all  training sets, are shown in figure 6. Likewise, figure 7 shows Dice scores generated by holding the data set size constant, and varying the level of augmentation. Only results from training datasets of sizes 1, 5 and 9 are shown.
\begin{table}
\centering
\caption{Mean Dice score over validation data, for different models}
\begin{tabular}{|l|l|l|l|l|l|}
\hline
\#Training subjects & 1     & 3     & 5     & 7     & 9     \\ \hline
Augmentation Level  &       &       &       &       &       \\ \hline
5x                  & 0.641 & 0.773 & 0.787 & 0.814 & 0.838 \\
10x                 & 0.682 & 0.798 & 0.802 & 0.832 & 0.850 \\
20x                 & 0.710 & 0.804 & 0.819 & 0.842 & 0.855 \\
30x                 & 0.721 & 0.812 & 0.824 & 0.851 & 0.860 \\
40x                 & 0.723 & 0.818 & 0.822 & 0.853 & 0.858 \\
50x                 & 0.729 & 0.817 & 0.826 & 0.861 & 0.863 \\ \hline
\end{tabular}
\end{table}

\begin{figure}
      \subfloat[Original Image]{\includegraphics[width=0.45\linewidth]{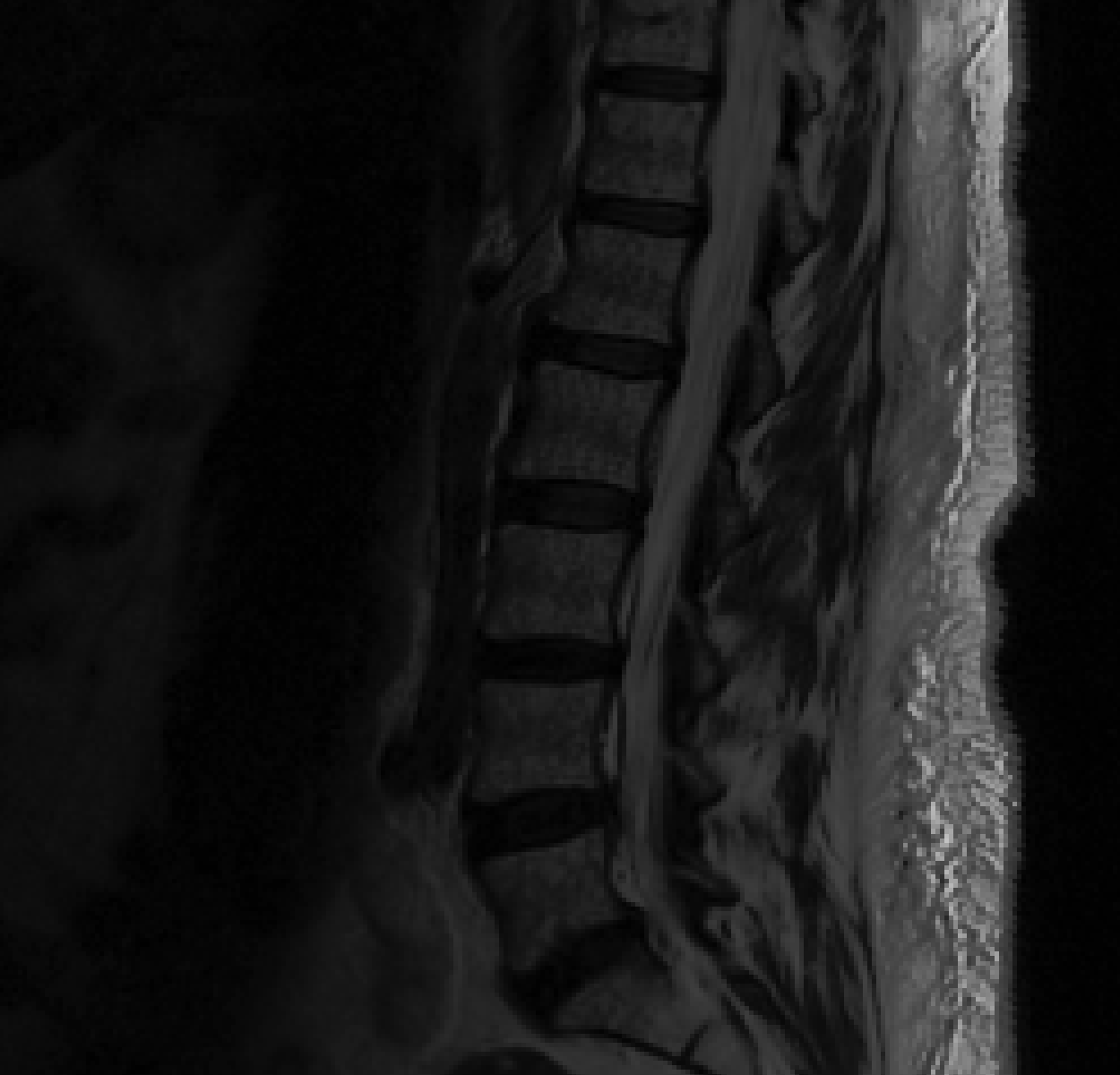}}
      \hfill{}
      \subfloat[Manual Segmentation]{\includegraphics[width=0.45\linewidth]{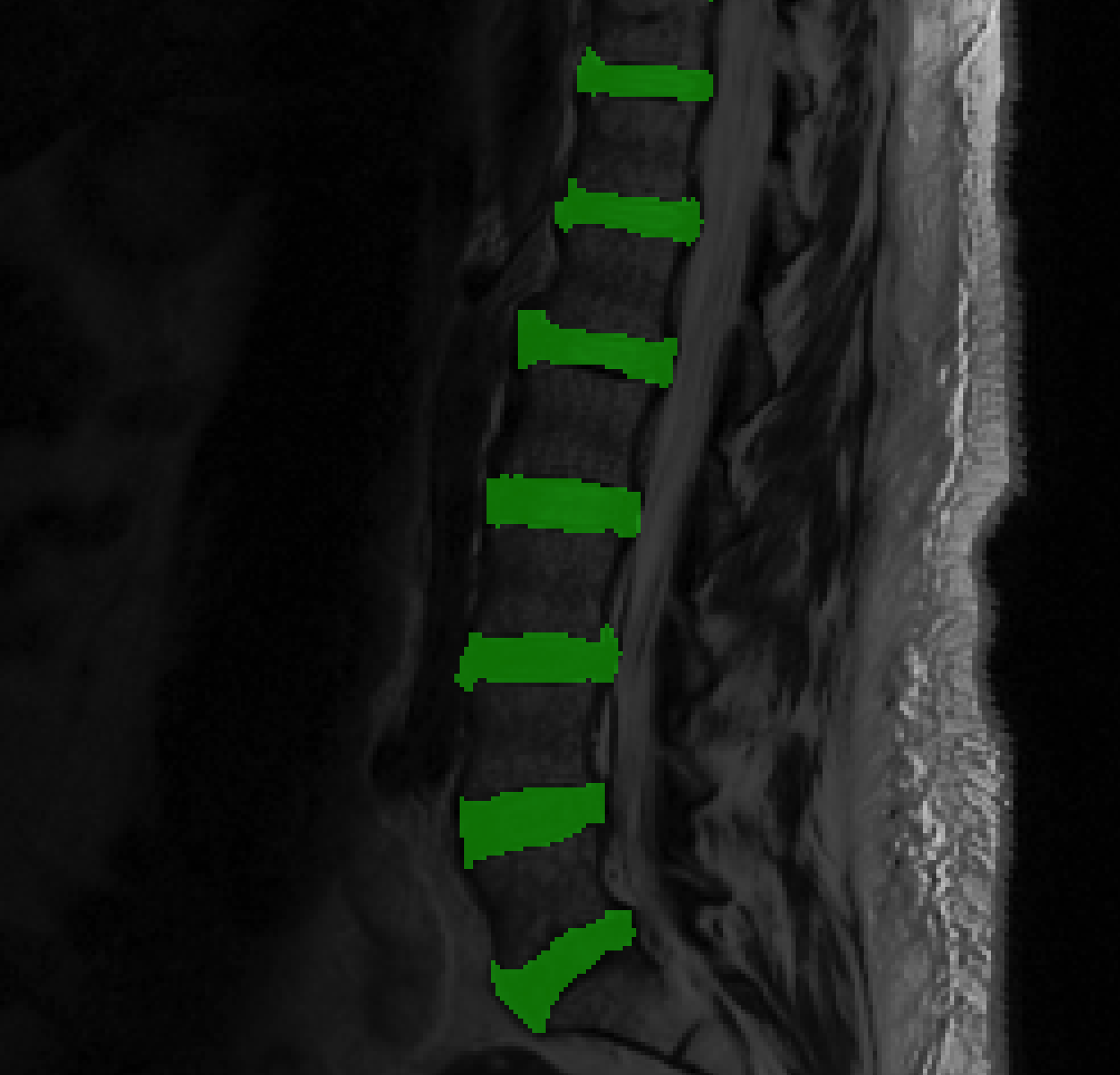}}
      \newline
      \subfloat[Single subject training]{\includegraphics[width=0.45\linewidth]{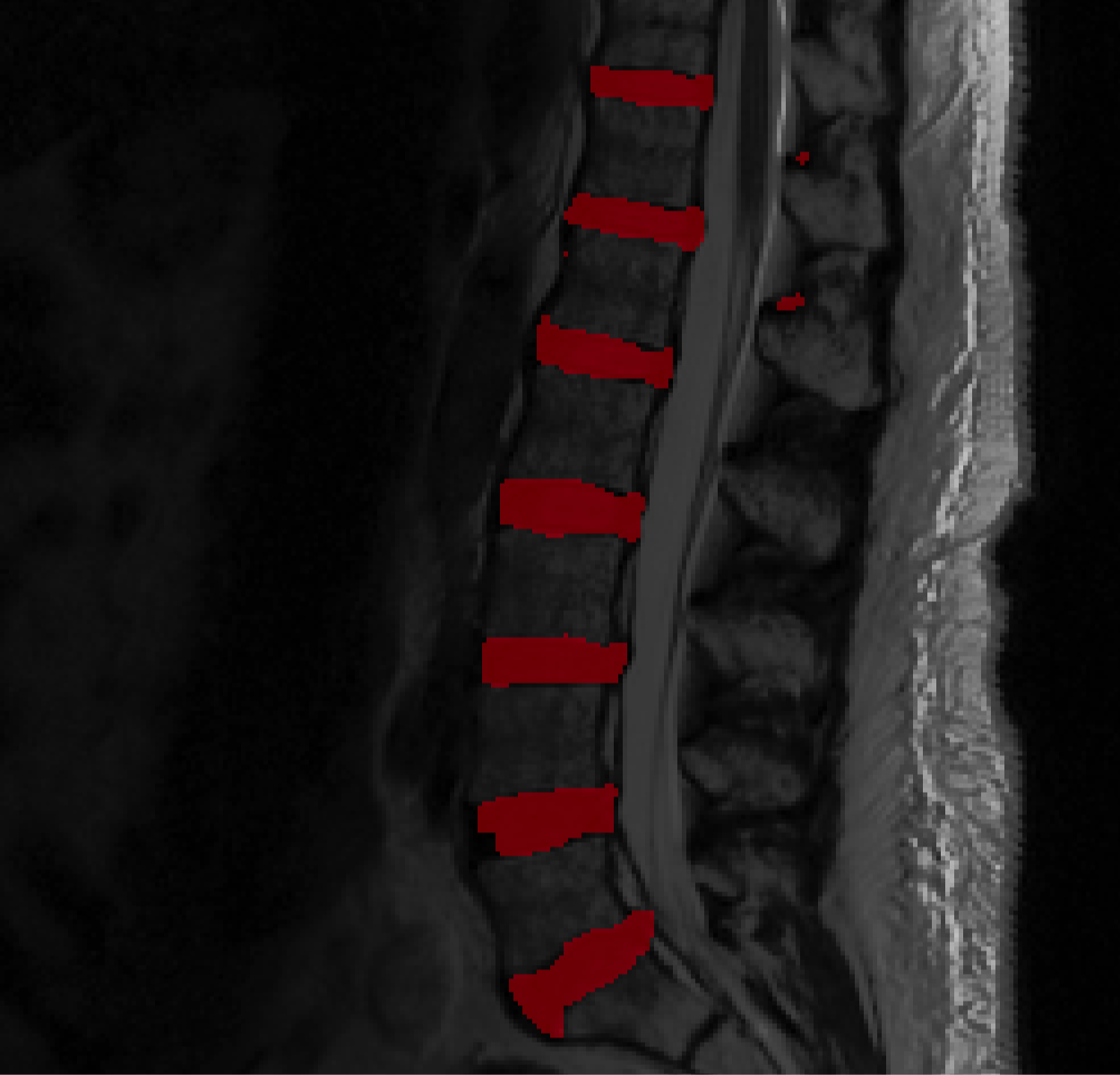}}
      \hfill
      \subfloat[Training with 9 subjects]{\includegraphics[width=0.45\linewidth]{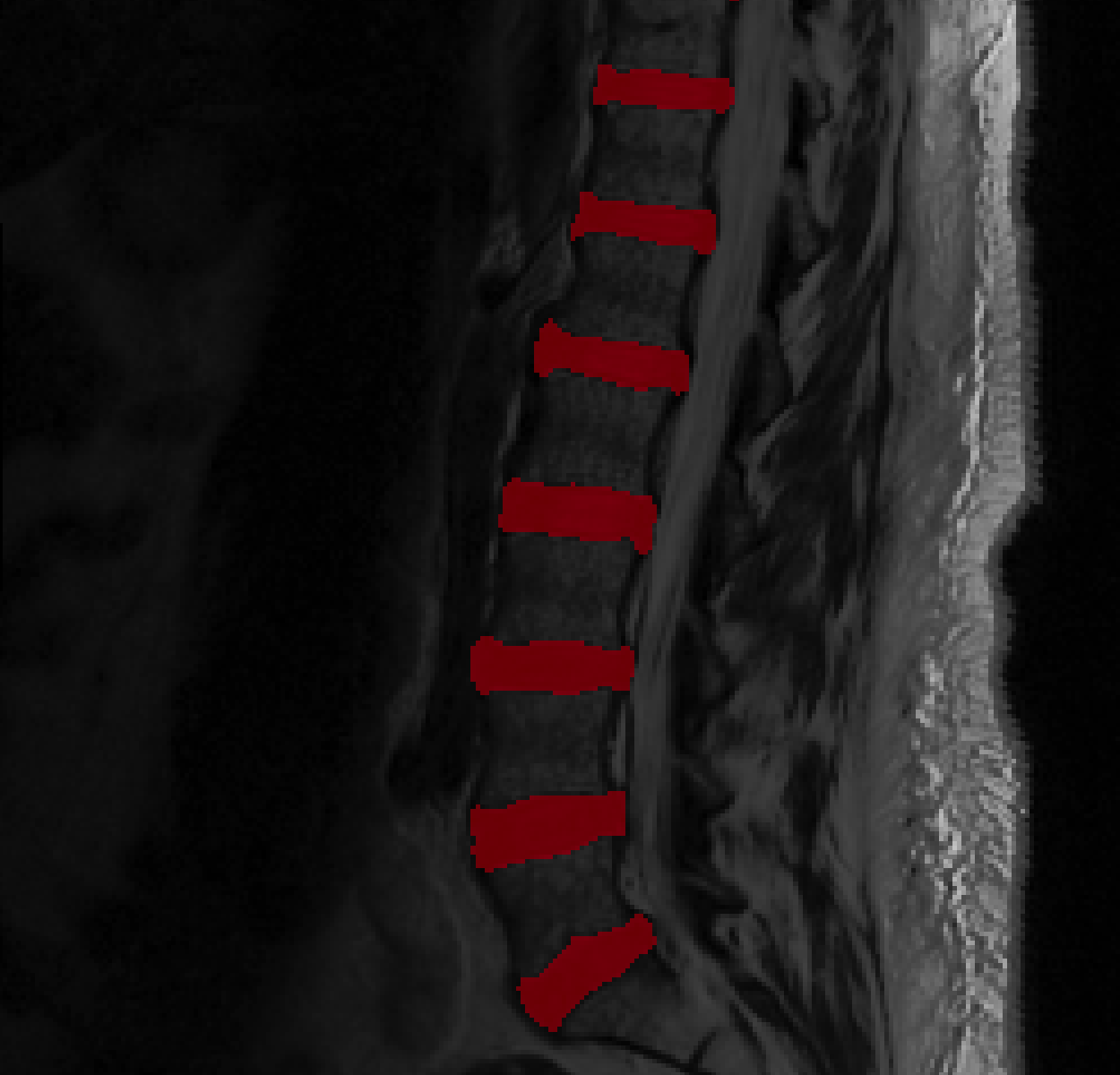}}
      \caption{Disk segmentation obtained using 50x augmented data (case picked from validation data set)}
      \label{fig:qual}
\end{figure}

\begin{figure}
  \centering
  \includegraphics[width=\linewidth]{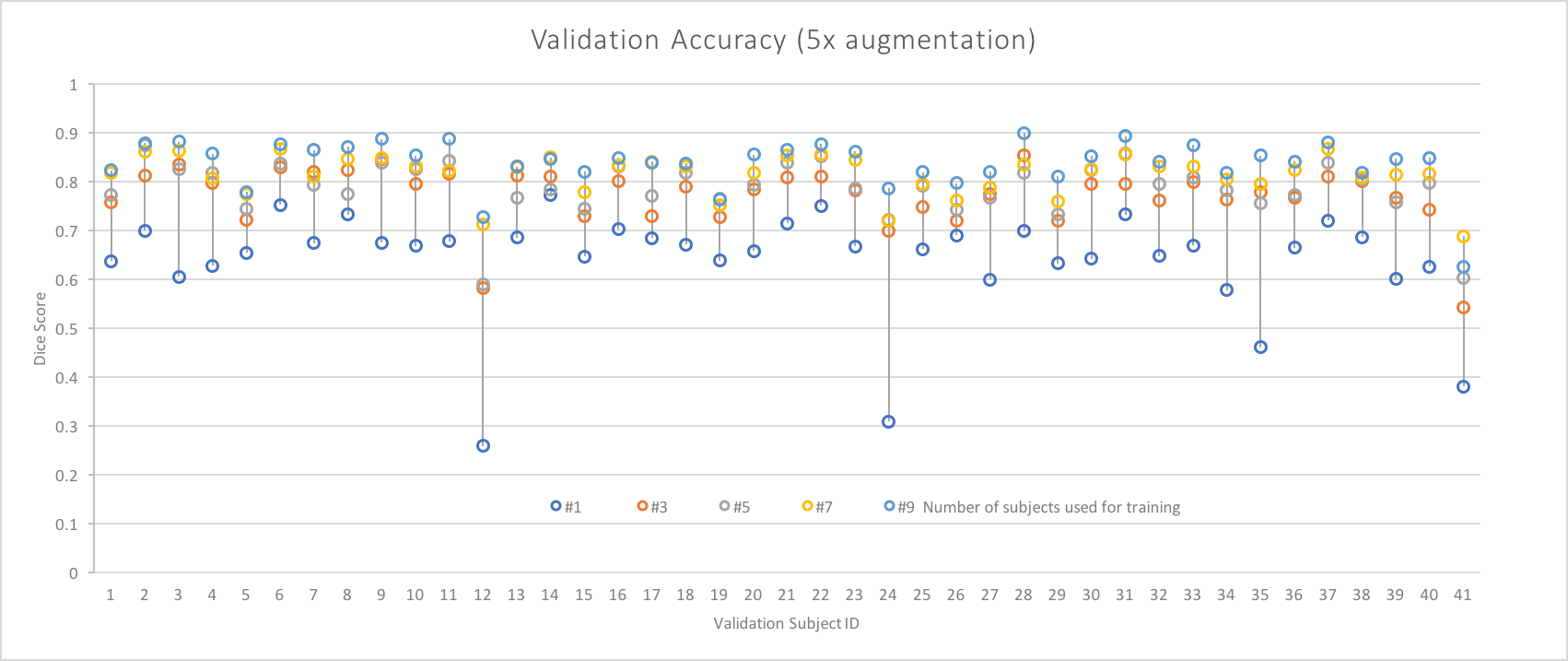}
  \includegraphics[width=\linewidth]{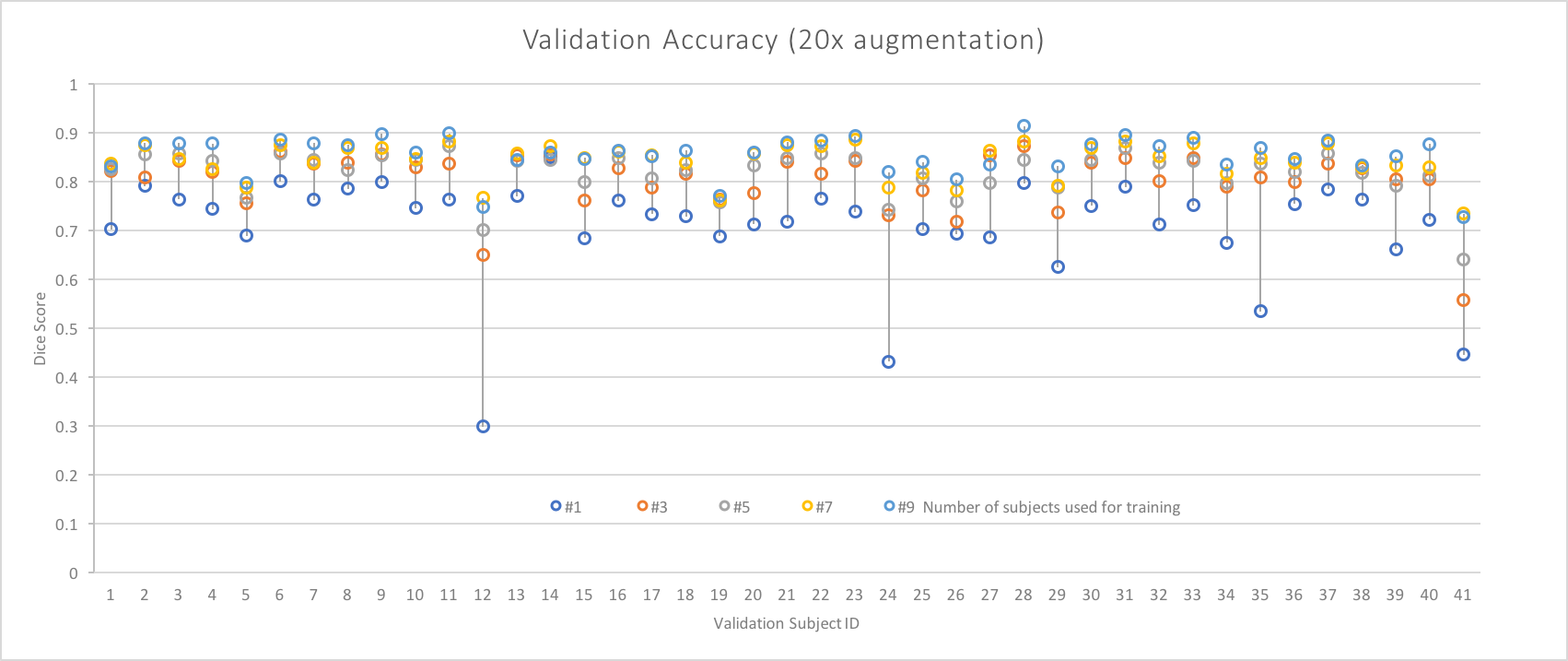}
  \includegraphics[width=\linewidth]{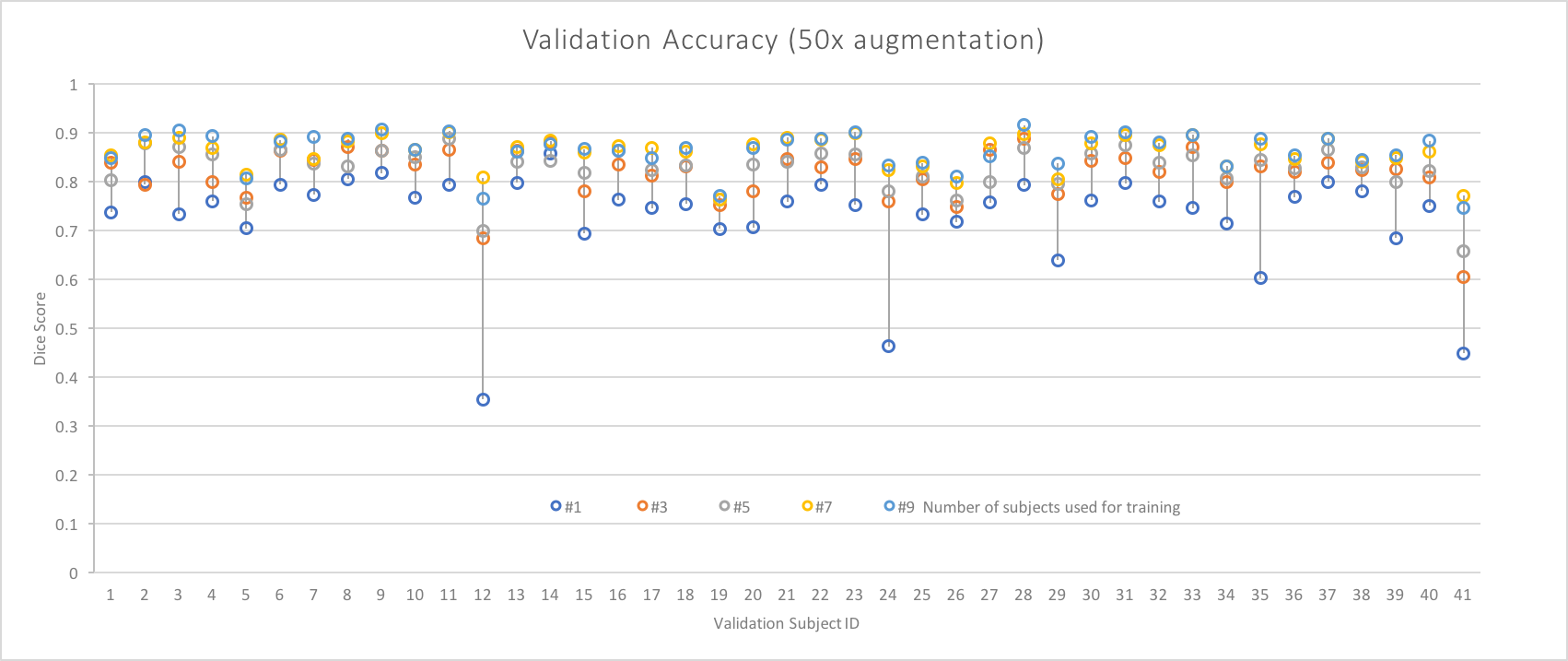}
  \caption{Validation Dice scores curves grouped by augmentation levels}
  \label{fig:valtrain}
\end{figure}

\begin{figure}
  \centering
  \includegraphics[width=\linewidth]{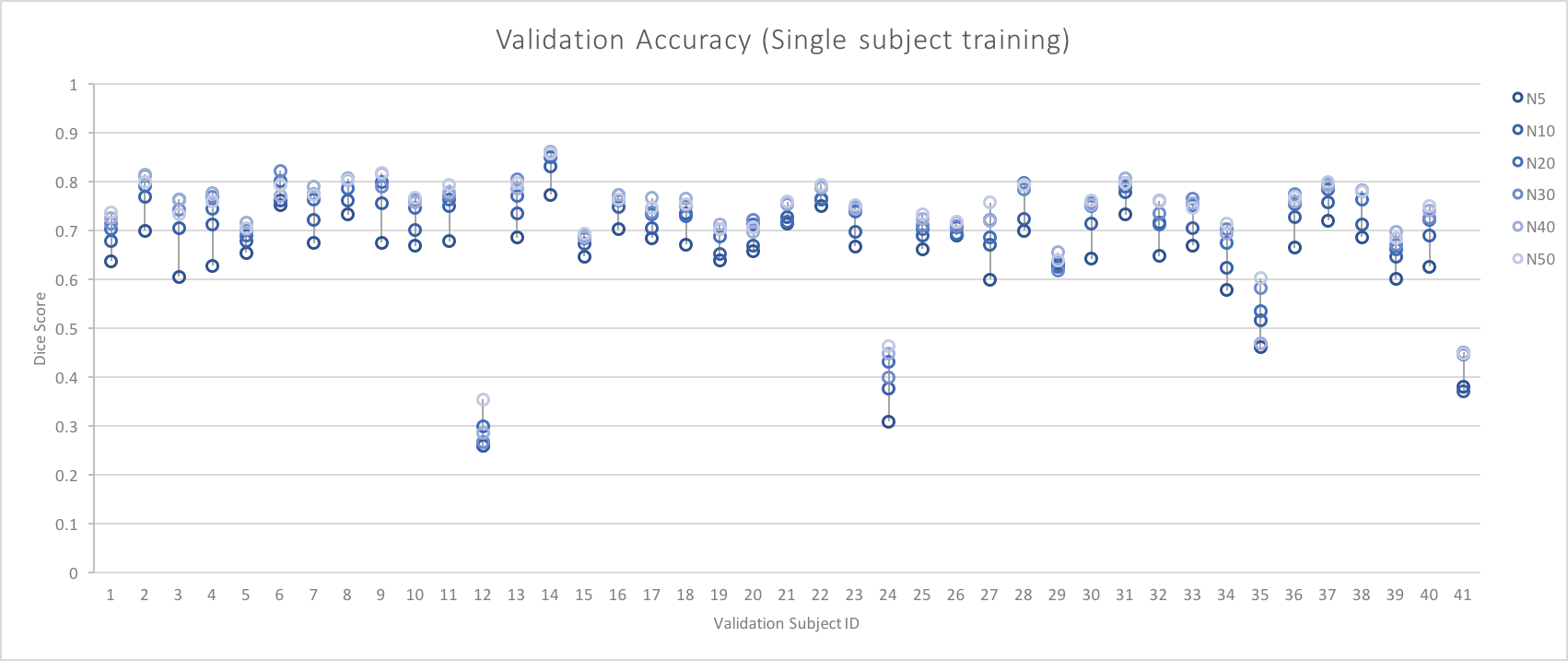}
  \includegraphics[width=\linewidth]{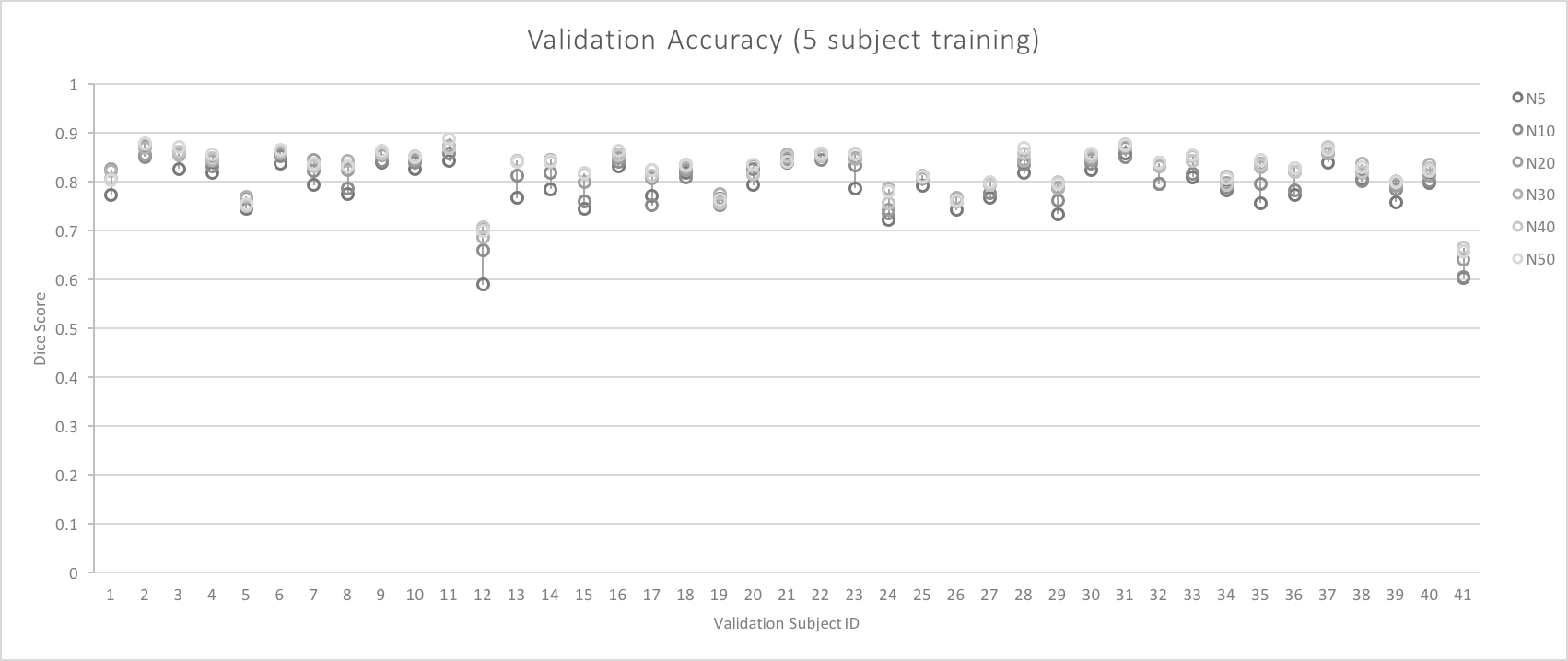}
  \includegraphics[width=\linewidth]{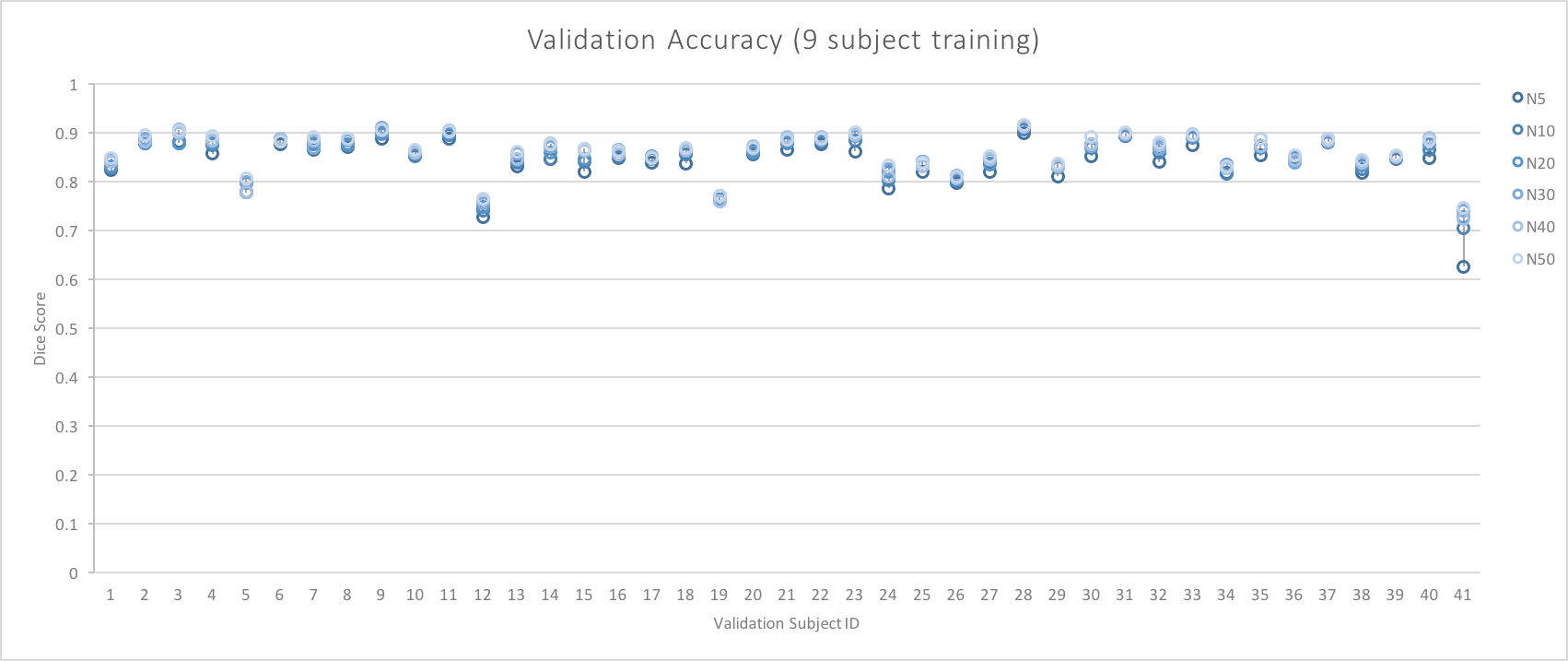}
  \caption{Validation Dice scores curves grouped by training data sizes}
  \label{fig:valaug}
\end{figure}

\section{CONCLUSIONS}
Figures 3 and 4 along with table 1 indicate that more augmentation leads to lower training error, as does training with larger datasets. Figures 5 and 6 alongside Table 2, confirm this trend in the validation dataset. Thus, we may conclude that, for a given training dataset, a greater  degree of augmentation leads to improved generalization accuracy.  This is expected, given the long history of augmentation. What is surprising is how well augmentation can work. The top row of figure 3 (and figure 6) show that at 5x augmentation our trained model struggles to `learn' how to segment disks. Bottom rows of the same figures show that with 50x augmentation, the situation is largely rectified. Perhaps, the most startling conclusion here is that a single scan, segmented by a human expert can be used by a deep machine learning algorithm to `learn' how to segment inter-vertebral disks. Perhaps, this is true of other anatomical regions of interest as well. Of course,  heavy augmentation (50x) is necessary to achieve this task (Fig 3, row 3). Figure 5 (c) shows that segmentation generated using a residual U-Net trained on a test case. While the segmentation in 5(c) is of poorer quality than that in 5(d) due to the presence of small noisy blobs in posterior to the spinal cord, it is still remarkable, given that the network was trained from exactly one delineated `scan'. All the disks are correctly localized and the delineation is comparable to human generated segmentation. This demonstrates the power of augmentation in training deep learning  segmentation models. 

Our experiments demonstrate that data augmentation improves both training and validation accuracy across the board, for a given training dataset size. This can be seen from figures 4 and 7. While the increase in accuracy due to augmentation is much higher in  single subject based training, an accuracy boost due to increased augmentation persists even when 9 separate scans are used for training. While this is expected, it underlines the fact that a properly designed augmentation strategy might be just as important as the structure of the neural network model being used. This is especially relevant given that both the computer vision and medical image analysis communities tend to be more focused on network architecture design. In practical applications for clinical usage, augmentation might be just as important.

Inter-vertebral disc segmentation is one of the fundamental tasks in spine image analysis. Dice scores in the range of 0.9+ have been reported in literature for this task. Yet, we wish to emphasize that we are  presenting results using small data sets in this paper. Even 9 scans is an extremely small dataset. Even in the pre-deep learning era, medical image segmentation algorithms were frequently trained on many more than `9' manually segmented cases. Thus, a  Dice score of 0.86 is quite impressive, especially since we are training a deep model with millions of tunable parameters and validating on a test set drawn from multiple scanners. With 50x augmentation and 9 scans for training, the model achieves a Dice score of 0.86. This is surprisingly high, given the limited size of training data. The fact that a small dataset can yield accurate segmentations opens the door to greatly accelarating the creation and usage of  deep learning based image segmentation algorithms in clinics around the world. The general wisdom in the medical imaging community, has been to collect a large manually annotated dataset prior to initiating a training process.  This is inspired by efforts like ImageNet \cite{deng2009imagenet} and COCO \cite{lin2014microsoft} in the computer vision community.  However, in medical imaging, collecting such large training datasets can be prohibitively expensive. This is especially true when a small clinic is attempting to use image segmentation to quantify a niche disease process. Extreme augmentation can be deployed in these use cases. It is relatively cheap to have a physician accurately delineate anatomy on a few scans. These scans can be used as training data with heavy augmentation to automatically segment a greater number of scans. Automatically segmented scans which are the output of this process can then be corrected by physicians and ploughed back into the training dataset. This process can scale much faster than having physicians delineate every single scan in a large dataset. Since physician time is scarce, an extreme augmentation based scheme for creating training datasets can be used to create larger training datasets far quicker than direct segmentation by clinicians.

\section{Discussion}
Work presented here reiterates how powerful data augmentation really is as a pre-processing step. The dominant research paradigm of in machine learning remains the creation and evaluation of newer learning models. Yet, effort directed towards improving and optimizing data augmentation can yield tremendous benefits. Perhaps, these benefits will accrue most heavily in communities where annotated data is limited and prohibitively expensive to obtain.

This study just scratches the surface, in regards to how best one can maximize the benefit that can be derived from augmentation. Yet, it's an important first step, that is especially relevant to the medical image analysis community. Perhaps there exist other symmetries in imaging, apart from rotation, scaling, shear and mirroring, which can be exploited for training better neural network models. Future studies must explore this topic in detail.

Lastly, we want to conclude this manuscript by re-stating that pre-processing imaging data using heavy augmentation is an easily implementable heuristic, which can substantially boost the performance of a trained deep learning model while reducing the need for large manually labelled data sets. While hyper-parameters associated with network architecture are generally favorite targets of study in model optimization, perhaps parameters associated with data augmentation should be given an equal weight in future studies.

\addtolength{\textheight}{-12cm}   



\section*{Acknowledgements}
We thank the UCLA and the NVIDIA corporation for supporting the studies presented here.


References are important to the reader; therefore, each citation must be complete and correct. If at all possible, references should be commonly available publications.

\bibliographystyle{IEEETrans.bst}
\bibliography{IEEE.bib}

\begin{thebibliography}{10}
\providecommand{\url}[1]{#1}
\csname url@samestyle\endcsname
\providecommand{\newblock}{\relax}
\providecommand{\bibinfo}[2]{#2}
\providecommand{\BIBentrySTDinterwordspacing}{\spaceskip=0pt\relax}
\providecommand{\BIBentryALTinterwordstretchfactor}{4}
\providecommand{\BIBentryALTinterwordspacing}{\spaceskip=\fontdimen2\font plus
\BIBentryALTinterwordstretchfactor\fontdimen3\font minus
  \fontdimen4\font\relax}
\providecommand{\BIBforeignlanguage}[2]{{%
\expandafter\ifx\csname l@#1\endcsname\relax
\typeout{** WARNING: IEEEtranS.bst: No hyphenation pattern has been}%
\typeout{** loaded for the language `#1'. Using the pattern for}%
\typeout{** the default language instead.}%
\else
\language=\csname l@#1\endcsname
\fi
#2}}
\providecommand{\BIBdecl}{\relax}
\BIBdecl

\bibitem{abadi2016tensorflow}
M.~Abadi, P.~Barham, J.~Chen, Z.~Chen, A.~Davis, J.~Dean, M.~Devin,
  S.~Ghemawat, G.~Irving, M.~Isard \emph{et~al.}, ``Tensorflow: a system for
  large-scale machine learning.'' in \emph{OSDI}, vol.~16, 2016, pp. 265--283.

\bibitem{chollet2015keras}
F.~Chollet \emph{et~al.}, ``Keras,'' 2015.

\bibitem{deng2009imagenet}
J.~Deng, W.~Dong, R.~Socher, L.-J. Li, K.~Li, and L.~Fei-Fei, ``Imagenet: A
  large-scale hierarchical image database,'' in \emph{2009 IEEE conference on
  computer vision and pattern recognition}.\hskip 1em plus 0.5em minus
  0.4em\relax Ieee, 2009, pp. 248--255.

\bibitem{Gaonkar2017}
B.~Gaonkar, Y.~Xia, D.~Villaroman, A.~Ko, M.~Attiah, J.~Beckett, and
  L.~Macyszyn, ``{Multi-Parameter Ensemble Learning for Automated Vertebral
  Body Segmentation in Heterogeneously Acquired Clinical MR Images},''
  \emph{IEEE Journal of Translational Engineering in Health and Medicine},
  vol.~5, 2017.

\bibitem{greenspan2016guest}
H.~Greenspan, B.~Van~Ginneken, and R.~M. Summers, ``Guest editorial deep
  learning in medical imaging: Overview and future promise of an exciting new
  technique,'' \emph{IEEE Transactions on Medical Imaging}, vol.~35, no.~5, pp.
  1153--1159, 2016.

\bibitem{han2018spine}
Z.~Han, B.~Wei, A.~Mercado, S.~Leung, and S.~Li, ``Spine-gan: Semantic
  segmentation of multiple spinal structures,'' \emph{Medical image analysis},
  2018.

\bibitem{He2017}
\BIBentryALTinterwordspacing
K.~He, ``{Mask R-CNN},'' 2017. [Online]. Available:
  \url{http://arxiv.org/abs/1703.06870{\%}5Cnhttps://arxiv.org/pdf/1703.06870.pdf}
\BIBentrySTDinterwordspacing

\bibitem{Jamaludin2016a}
A.~Jamaludin, T.~Kadir, and A.~Zisserman, ``{SpineNet: Automatically
  pinpointing classification evidence in spinal MRIs},'' in \emph{Lecture Notes
  in Computer Science (including subseries Lecture Notes in Artificial
  Intelligence and Lecture Notes in Bioinformatics)}, vol. 9901 LNCS, 2016, pp.
  166--175.

\bibitem{Jamaludin2016}
A.~Jamaludin, M.~Lootus, T.~Kadir, and A.~Zisserman, ``{Automatic
  intervertebral discs localization and segmentation: A vertebral approach},''
  in \emph{Lecture Notes in Computer Science (including subseries Lecture Notes
  in Artificial Intelligence and Lecture Notes in Bioinformatics)}, vol. 9402,
  2016, pp. 97--103.

\bibitem{Jamaludin2017}
A.~Jamaludin, M.~Lootus, T.~Kadir, A.~Zisserman, J.~Urban, M.~C. Batti{\'{e}},
  J.~Fairbank, and I.~McCall, ``{ISSLS PRIZE IN BIOENGINEERING SCIENCE 2017:
  Automation of reading of radiological features from magnetic resonance images
  (MRIs) of the lumbar spine without human intervention is comparable with an
  expert radiologist},'' \emph{European Spine Journal}, vol.~26, no.~5, pp.
  1374--1383, 2017.

\bibitem{Kazemi2014}
V.~Kazemi and J.~Sullivan, ``{One millisecond face alignment with an ensemble
  of regression trees},'' in \emph{Proceedings of the IEEE Computer Society
  Conference on Computer Vision and Pattern Recognition}, 2014, pp. 1867--1874.

\bibitem{krizhevsky2012imagenet}
A.~Krizhevsky, I.~Sutskever, and G.~E. Hinton, ``Imagenet classification with
  deep convolutional neural networks,'' in \emph{Advances in neural information
  processing systems}, 2012, pp. 1097--1105.

\bibitem{Larobina2014}
M.~Larobina and L.~Murino, ``{Medical image file formats},'' pp. 200--206,
  2014.

\bibitem{Li:2016kj}
X.~Li, P.~S. Morgan, J.~Ashburner, J.~Smith, and C.~Rorden, ``{The first step
  for neuroimaging data analysis: DICOM to NIfTI conversion},'' \emph{Journal
  of Neuroscience Methods}, vol. 264, pp. 47--56, may 2016.

\bibitem{lin2014microsoft}
T.-Y. Lin, M.~Maire, S.~Belongie, J.~Hays, P.~Perona, D.~Ramanan,
  P.~Doll{\'a}r, and C.~L. Zitnick, ``Microsoft coco: Common objects in
  context,'' in \emph{European conference on computer vision}.\hskip 1em plus
  0.5em minus 0.4em\relax Springer, 2014, pp. 740--755.

\bibitem{Lootus2015}
M.~Lootus, T.~Kadir, and A.~Zisserman, ``{Automated radiological grading of
  spinal MRI},'' \emph{Lecture Notes in Computational Vision and Biomechanics},
  vol.~20, pp. 119--130, 2015.

\bibitem{lowekamp2013design}
B.~C. Lowekamp, D.~T. Chen, L.~Ib{\'a}{\~n}ez, and D.~Blezek, ``The design of
  simpleitk,'' \emph{Frontiers in neuroinformatics}, vol.~7, p.~45, 2013.

\bibitem{Moran2018}
S.~Moran, B.~Gaonkar, W.~Whitehead, A.~Wolk, L.~Macyszyn, and S.~Iyer, ``{Deep
  learning for medical image segmentation-using the IBM TrueNorth neurosynaptic
  system},'' in \emph{Progress in Biomedical Optics and Imaging - Proceedings
  of SPIE}, vol. 10579, 2018.

\bibitem{Nair:2010vq}
V.~Nair and G.~E. Hinton, ``{Rectified linear units improve restricted
  boltzmann machines},'' in \emph{Proceedings of the 27th international
  {\{}{\ldots}{\}}}, 2010.

\bibitem{Razavian2014}
A.~S. Razavian, H.~Azizpour, J.~Sullivan, and S.~Carlsson, ``{CNN features
  off-the-shelf: An astounding baseline for recognition},'' in \emph{IEEE
  Computer Society Conference on Computer Vision and Pattern Recognition
  Workshops}, 2014, pp. 512--519.

\bibitem{Ronneberger:2015gk}
O.~Ronneberger, P.~Fischer, and T.~Brox, ``{U-Net: Convolutional Networks for
  Biomedical Image Segmentation},'' in \emph{Medical Image Computing and
  Computer-Assisted Intervention {\{}$\backslash$textendash{\}} MICCAI
  2015}.\hskip 1em plus 0.5em minus 0.4em\relax Cham: Springer, Cham, oct 2015,
  pp. 234--241.

\bibitem{Sekuboyina2017}
\BIBentryALTinterwordspacing
A.~Sekuboyina, A.~Valentinitsch, J.~S. Kirschke, and B.~H. Menze, ``{A
  Localisation-Segmentation Approach for Multi-label Annotation of Lumbar
  Vertebrae using Deep Nets},'' no.~1, pp. 1--10, 2017. [Online]. Available:
  \url{http://arxiv.org/abs/1703.04347}
\BIBentrySTDinterwordspacing

\bibitem{shaikhina2017handling}
T.~Shaikhina and N.~A. Khovanova, ``Handling limited datasets with neural
  networks in medical applications: A small-data approach,'' \emph{Artificial
  intelligence in medicine}, vol.~75, pp. 51--63, 2017.

\bibitem{Terminology1970}
C.~P. Terminology, ``{Current procedural terminology (CPT).}'' \emph{JAMA : the
  journal of the American Medical Association}, vol. 212, pp. 873--874, 1970.

\bibitem{wachinger2018deepnat}
C.~Wachinger, M.~Reuter, and T.~Klein, ``Deepnat: Deep convolutional neural
  network for segmenting neuroanatomy,'' \emph{NeuroImage}, vol. 170, pp.
  434--445, 2018.

\bibitem{Warfield2004}
S.~K. Warfield, K.~H. Zou, and W.~M. Wells, ``{Simultaneous truth and
  performance level estimation (STAPLE): An algorithm for the validation of
  image segmentation},'' \emph{IEEE Transactions on Medical Imaging}, vol.~23,
  no.~7, pp. 903--921, 2004.

\bibitem{Whitehead2018}
W.~Whitehead, S.~Moran, B.~Gaonkar, L.~Macyszyn, and S.~Iyer, ``{A deep
  learning approach to spine segmentation using a feed-forward chain of
  pixel-wise convolutional networks},'' in \emph{Proceedings - International
  Symposium on Biomedical Imaging}, vol. 2018-April, 2018.

\bibitem{witten2016data}
I.~H. Witten, E.~Frank, M.~A. Hall, and C.~J. Pal, \emph{Data Mining: Practical
  machine learning tools and techniques}.\hskip 1em plus 0.5em minus
  0.4em\relax Morgan Kaufmann, 2016.

\bibitem{Yushkevich:2006fk}
P.~A. Yushkevich, J.~Piven, H.~C. Hazlett, R.~G. Smith, S.~Ho, J.~C. Gee, and
  G.~Gerig, ``{User-guided 3D active contour segmentation of anatomical
  structures: Significantly improved efficiency and reliability},''
  \emph{Neuroimage}, vol.~31, no.~3, pp. 1116--1128, jul 2006.

\bibitem{ResUNet}
\BIBentryALTinterwordspacing
Z.~Zhang, Q.~Liu, and Y.~Wang, ``Road extraction by deep residual u-net,''
  \emph{CoRR}, vol. abs/1711.10684, 2017. [Online]. Available:
  \url{http://arxiv.org/abs/1711.10684}
\BIBentrySTDinterwordspacing

\bibitem{Zheng2017}
G.~Zheng, C.~Chu, D.~L. Belav{\'{y}}, B.~Ibragimov, R.~Korez, T.~Vrtovec,
  H.~Hutt, R.~Everson, J.~Meakin, I.~L. Andrade, B.~Glocker, H.~Chen, Q.~Dou,
  P.~A. Heng, C.~Wang, D.~Forsberg, A.~Neubert, J.~Fripp, M.~Urschler,
  D.~Stern, M.~Wimmer, A.~A. Novikov, H.~Cheng, G.~Armbrecht, D.~Felsenberg,
  and S.~Li, ``{Evaluation and comparison of 3D intervertebral disc
  localization and segmentation methods for 3D T2 MR data: A grand
  challenge},'' \emph{Medical Image Analysis}, vol.~35, pp. 327--344, 2017.

\bibitem{Zou2004}
K.~H. Zou, S.~K. Warfield, A.~Bharatha, C.~M. Tempany, M.~R. Kaus, S.~J. Haker,
  W.~M. Wells, F.~A. Jolesz, and R.~Kikinis, ``{Statistical Validation of Image
  Segmentation Quality Based on a Spatial Overlap Index},'' \emph{Academic
  Radiology}, vol.~11, no.~2, pp. 178--189, 2004.

\end{thebibliography}

\end{document}